\documentclass{article}
\usepackage{amsmath}
\usepackage{amsfonts}
\usepackage{dsfont}
\usepackage{url}
\usepackage{color}
\usepackage{graphicx}
\usepackage{float}
\usepackage[export]{adjustbox}
\graphicspath{{../images/}}

\usepackage{mlspconf}

\newcommand\given[1][]{\:#1\vert\:}

\title{Robust importance-weighted cross-validation\\ under sample selection bias}
\name{Wouter M. Kouw $^{1,3}$\sthanks{Majority of work done while WK was at Delft University of Technology.}, Jesse H. Krijthe $^{2}$, Marco Loog $^{3}$}
\address{$^{1}$ Department of Electrical Engineering, Eindhoven University of Technology\\ 
$^{2}$ Institute for Computing and Information Sciences, Radboud University Nijmegen\\ 
$^{3}$ Department of Intelligent Systems, Delft University of Technology}

\begin{document}
%

\maketitle
\begin{abstract}
Cross-validation under sample selection bias can, in principle, be done by importance-weighting the empirical risk. However, the importance-weighted risk estimator produces sub-optimal hyperparameter estimates in problem settings where large weights arise with high probability. We study its sampling variance as a function of the training data distribution and introduce a control variate to increase its robustness to problematically large weights.
\end{abstract}
\begin{keywords}
Sample selection bias, cross-validation.
\end{keywords}
%
%

\section{Introduction}
Classification under sample selection bias refers to settings where training data is collected locally, but the goal is to generalize to a larger target population \cite{zadrozny2004learning,cortes2008sample,quionero2009dataset}. For example, data collected in one hospital with the aim of generalizing to a national population or data collected from a process that is assumed to be stationary, but in reality drifts slowly over time. To control for selection bias, each sample is weighted by a factor that matches its observation probability in the training data distribution to that in the target population, a process known as \emph{importance-weighting} \cite{sugiyama2007covariate,cortes2008sample}. However, importance-weighting underperforms and can even fail, if the weights are too large \cite{cortes2008sample}. The weights directly scale the estimator's sampling variance (see Section \ref{sec:sampling_var}) and sampling skewness \cite{kouw2018effects}. Importance-weighting is therefore not suited to problem settings where large weights occur with high probability. It needs to be more robust to be considered a practical solution.

One could consider weight truncation as a means of avoiding large weights, but that introduces a substantial bias in the estimator \cite{ionides2008truncated}. Alternatively, variance reduction techniques incorporate additional information about the statistic of interest to avoid introducing a bias. Control variates are such a technique, requiring an additional function of the random variable that correlates well with the statistic of interest and whose expected value is known \cite{nelson1987control}. Because of the correlation, the statistic is known to rise -- or fall in case of negative correlation -- above its expected value whenever the control variate does. The deviation of the control variate from its expected value can be subtracted from the statistic, thereby reducing the statistic's deviation from its expectation (i.e. sampling variance) \cite{nelson1987control,mcbook}. In importance-weighting, the importance weights themselves constitute an additional function that correlates with the importance-weighted loss and whose expected value is known \cite{owen2000safe}. An importance-weighted control variate can reduce the sampling variance of an importance-weighted risk estimator, thereby increasing its robustness to large weights.

We are interested in cross-validation under sample selection bias. In standard cross-validation, the risk is repeatedly estimated and averaged to obtain an estimate of the expected generalization error \cite{kohavi1995study}. With this estimate, optimal hyperparameters can be selected. Cross-validation becomes difficult under sample selection bias, as the estimator essentially over-fits to the distribution from which the training data is generated \cite{cawley2010over}. Importance-weighted cross-validation has been explored as a solution \cite{sugiyama2007covariate}, but is known to produce sub-optimal hyperparameter estimates in problem settings with large weights \cite{kouw2018effects}. We argue for including the weight-based control variate in the importance-weighted risk estimator such that importance-weighted cross-validation is more robust to large weights.

Our contributions can be summarized as follows:
\begin{itemize}
\item We apply a controlled importance-weighted risk estimator to a cross-validation procedure.
\item We empirically show that the inclusion of the control variate increases robustness to large weights.
\end{itemize}
In the remainder of the paper we present the problem setting in more detail (Section \ref{sec:covshift}), discuss the importance-weighted risk estimator (Section \ref{sec:iw}) and show how the control variate reduces sampling variance (Section \ref{sec:control}). We then run a cross-validation experiment to demonstrate the effectiveness of the controlled estimator (Section \ref{sec:cross}). We briefly discuss conclusions and extensions in Section \ref{sec:discussion}. 

\section{Sample selection bias} \label{sec:covshift}
Consider an input space ${\cal X} \subseteq \mathbb{R}^{D}$, and an output space ${\cal Y}$, with ${\cal Y} = \{-1, +1\}$ for binary classification or ${\cal Y} \subseteq \mathbb{N}$ for multi-class. The population, or \emph{target domain}, is a probability distribution $p_{\cal T}$ defined over this pair of spaces. The distribution of the training data $p_{\cal S}$, gathered under sample selection bias, is referred to as the \emph{source domain}. Under sample selection bias, the data distributions differ $p_{\cal S}(x) \neq p_{\cal T}(x)$ with $p_{\cal S}(x)$ having a smaller variance than $p_{\cal T}(x)$, but the posterior distributions remain equal $p_{\cal S}(y\given x) = p_{\cal T}(y\given x)$. Data points from the source domain are referred to as $x_i$ with labels $y_i$, while target domain data is referred to as $z_j$ with labels $u_j$. The challenge is to use labeled source data $\{(x_i, y_i)\}_{i=1}^{n}$ and unlabeled target data $\{z_j\}_{j=1}^{m}$ to predict target labels $u_j$.

Consider a function $h$, parameterized by $\theta$, that maps a data point to a real number, $h_{\theta} : {\cal X} \rightarrow \mathbb{R}$. The real-valued outputs are predictions of classes and we therefore refer to $h$ as a \emph{classifier}. Predictions are evaluated using a loss function $\ell : \mathbb{R} \times {\cal Y} \rightarrow \mathbb{R}$, which compares the prediction to the true label. The expected loss is called the \emph{risk} of the classifier, $R(h_{\theta})$ and the average loss is called the empirical risk $\hat{R}(h_\theta)$.

\subsection{An example setting} \label{sec:example}
\begin{figure}[htb]
\centering
\includegraphics[width=.42\textwidth]{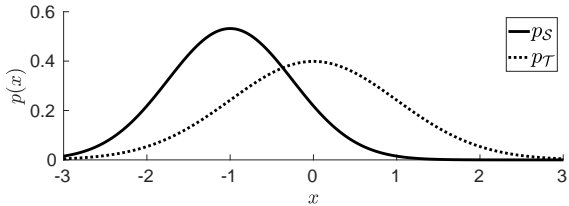}
\caption{Data distributions in the example setting.}
\label{fig:example}
\end{figure}

Throughout the paper, we illustrate the behavior of estimators with a running example of a classification problem under sample selection bias. The posterior distribution in both domains is a unit cumulative normal distribution, $p_{\cal T}(y\given x) = \Phi(y x \given 0, 1)$. The target data distribution is taken to be a unit normal distribution $p_{\cal T}(x) = \mathcal{N}(0, 1)$,  while the source data distribution is set to $\mathcal{N}(-1, \gamma)$. Figure \ref{fig:example} plots the data distributions, for $\gamma = 1/\sqrt{2}$. Note that $p_{\cal S}$ has higher probability mass between $x=-3$ and $x=-1/2$, implying these values will tend to be observed more often in the source data set than in the target set.

\noindent In order to evaluate risk, a choice of loss function and classifier is needed. For this example, we use a quadratic loss, $\ell(h_{\theta}( x),y) = \big(x \theta - y \big)^2$, with $\theta = 1/\sqrt{\pi}$ as coefficient. 

\section{Importance-weighting} \label{sec:iw}
Empirical risk minimization describes a classifier's performance by its expected loss. The risk function $R$ integrates the loss $\ell$ over a joint distribution $p$, and is therefore domain-specific. We are interested in minimizing the target risk $R_{\cal T}$:
\begin{align}
	R_{\cal T}(h_{\theta}) =& \int_{\cal X} \int_{\cal Y} \ell \big( h_{\theta} ( x ), y \big) p_{\cal T}\big(x,y \big) \ \mathrm{d}y \mathrm{d}x \, ,
\end{align}
which can be estimated with the sample average over data drawn from the target domain: 
\begin{align}
	\hat{R}_{\cal T}(h_{\theta}) = \frac{1}{m} \sum_{j=1}^{m} \ell \big( h_{\theta} \big( z_j \big) ,u_j \big) \, .
\end{align}
However, this estimator cannot be used, since we lack target labels $u$.

We are interested in estimators that do \emph{not} depend on $u$. One possibility is the importance-weighted risk estimator, which re-formulates the target risk to include the source distribution: $\iint \ell \big( h_{\theta} ( x ), y \big) p_{\cal S}(x,y) \big[p_{\cal T}(x,y) / p_{\cal S}(x,y) \big] \mathrm{d}y \mathrm{d}x $. Under sample selection bias, $p_{\cal T}(y \given x) = p_{\cal S}(y \given x)$, which simplifies the re-formulation to:
\begin{align}
	R_{\cal W}(h_{\theta}) =& \int_{\cal X} \int_{\cal Y} \ell \big(h_{\theta} \big( x \big),y \big) \frac{p_{\cal T}(x)}{p_{\cal S}(x)} p_{\cal S} \big( x, y \big) \ \mathrm{d}y \mathrm{d} x \label{eq:R_W} \, .
\end{align}
Using samples from the source domain $\{(x_i, y_i)\}_{i=1}^{n}$, this risk can be estimated through: 
\begin{align}
	\hat{R}_{\cal W}(h_{\theta}) = \frac{1}{n} \sum_{i=1}^{n} \ell( h_{\theta}(x_i), y_i) \frac{p_{\cal T}(x_i)}{p_{\cal S}(x_i)} \, . \label{eq:hatR_W}
\end{align} 
Note that it does not depend on target labels $u$. 

\subsection{Weights}
\begin{figure}[htb]
\centering
\includegraphics[width=.42\textwidth]{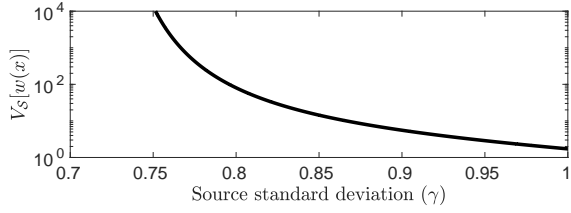}
\caption{Variance of the importance weights $w(x)$ as a function of the standard deviation $\gamma$ of the source data distribution in the example setting.}
\label{fig:weights}
\end{figure}

The ratio of probability distributions is referred to as the importance weight function $w(x) = p_{\cal T}(x) / p_{\cal S}(x)$. Each weight corrects the probability of observing a single pair $(x_i,y_i)$ under the source distribution to its probability under the target distribution. Higher values indicate higher importance to the target domain. In the example setting, the weight function is: $w(x) = \gamma \exp \big([\gamma^{-2}(-1 - x)^2 -x^2] / 2 \big)$. Its variance has an analytical solution, $\mathbb{V}_{\cal S}[ w(x) ] = \gamma^2 / (2\gamma^2-1) \cdot \exp \big(1/ (2\gamma^2 - 1) \big) - 1$, provided $\gamma > 1/\sqrt{2}$, shown in Figure \ref{fig:weights}. Note that the weights' variance grows exponentially as $\gamma$ shrinks. In other words, as the source domain becomes a more limited view of the target population, the spread over the importance of each sample grows sharply.

\subsection{Sampling variance} \label{sec:sampling_var}
In expectation, the estimators $\hat{R}_{\cal T}$ and $\hat{R}_{\cal W}$ are equivalent. However, they behave differently for finite sample sizes. The variance of an estimator with respect to data, is known as its \emph{sampling variance}. In the following, we will compare the sampling variances of $\hat{R}_{\cal W}$ and $\hat{R}_{\cal T}$.

Assuming data is drawn independently and is identically distributed (\emph{iid}), the sampling variance of the target risk estimator is
\begin{align}
\mathbb{V}_{\cal T} [ \hat{R}_{\cal T} ] 
	=& \ \mathbb{E}_{\cal T} \big[ \big( \frac{1}{m} \sum_{j=1}^{m} \ell( h_{\theta} ( z_j), u_j) - R_{\cal T} \big)^2 \big] \nonumber \\
	=& \ \frac{1}{m} \ \mathbb{E}_{\cal T} \big[ \big(\ell( h_{\theta}( x) ,y) - R_{\cal T} \big)^2 \big] = \frac{\sigma^2_{\cal T}}{m} \label{eq:sv02} \, ,
\end{align}
where $\sigma^2_{\cal T}$ is the sampling variance of $\hat{R}_{\cal T}$ with respect to a single point. Assuming the source data is drawn \emph{iid} as well, the sampling variance of the importance-weighted risk estimator follows similarly: 
\begin{align}
	\mathbb{V}_{\cal S} [ \hat{R}_{\cal W} ] 
	=& \ \mathbb{E}_{\cal S} \big[ \big(\frac{1}{n}\sum_{i=1}^n \ell(h_{\theta}(x_i), y_i) w(x_i) - R_{\cal W} \big)^2 \big] \nonumber \\
	=& \frac{1}{n} \ \mathbb{E}_{\cal S} \big[ \big( \ell( h_{\theta}( x ), y) w(x)- R_{\cal W} \big)^2 \big] = \frac{\sigma^2_{\cal W}}{n} \label{eq:sv03} \, .
\end{align}
Since $R_{\cal W} = R_{\cal T}$, Equations (\ref{eq:sv02}) and (\ref{eq:sv03}) reveal that the sampling variances of $\hat{R}_{\cal T}$ and $\hat{R}_{\cal W}$ only differ by a scaling induced by the importance weights:
\begin{align}
\sigma^2_{\cal T} - \sigma^2_{\cal W} 
=& \ \mathbb{E}_{\cal T} \big[ \ \ell( h_{\theta}( x ), y)^2 \big(1 - w(x) \big) \ \big] \, .
\end{align} 
Hence, in problem settings where large weights occur often, the importance-weighted risk estimator has a higher sampling variance than the target risk estimator. This makes sense: as the training data distribution provides a more narrow view of the population, the risk estimate becomes more uncertain.

\noindent Figure \ref{fig:sv02} plots $\sigma^2_{\cal W}$ and $\sigma^2_{\cal T}$ for the example setting ($\sigma^2_{\beta}$ will be discussed in Section \ref{sec:control}). Note that the shape of the curve of $\hat{R}_{\cal W}$ reflects the influence of weight variance (see Figure \ref{fig:weights}).

\subsection{Occurrence of large weights}
\begin{figure}[htb]
\includegraphics[width=.45\textwidth]{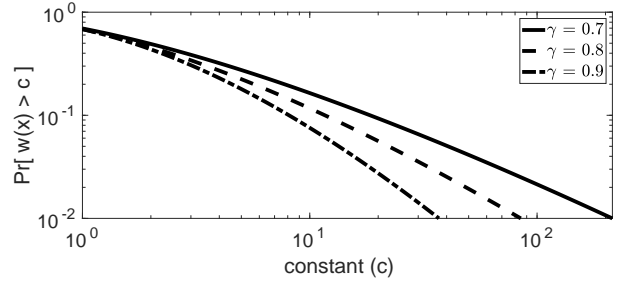}
\caption{Probability of an importance weight exceeding the constant $c$, for different values of source standard deviation $\gamma$.}
\label{fig:prob_wc}
\end{figure}

A large sampling variance does not necessarily mean a given data set will contain a large weight. Drawing a point with a large weight occurs with some probability. For illustration, we plot the probability that an importance-weight exceeds a constant $c$ in the example setting. Figure \ref{fig:prob_wc} shows curves for three choices of $\gamma$. For $\gamma = 0.7$, the probability of a weight exceeding the value $10$ is about $0.2$.

\section{Reducing sampling variance} \label{sec:control}

\subsection{Control variates}
A control variate is a function $g$ of a random variable $x$ that correlates with a particular statistic $f(x)$ of that random variable and whose expected value is known \cite{nelson1987control,mcbook}. It reduces sampling variance as follows: suppose that $g(x)$ is positively correlated to the statistic $f(x)$. Then, whenever $g(x_i)$ rises above its expected value, $g(x_i) - \mathbb{E}[g(x)] > 0$, $f(x_i)$ will also rise above its expected value, $f(x_i) - \mathbb{E}[f(x)] > 0$. If $g$ is negatively correlated, then $f$ will fall below its expectation whenever $g$ rises above its expectation. By subtracting the control variate from the statistic, $f(x_i) - (g(x_i) - \mathbb{E}[g(x)])$, the statistic's deviation from its expected value can be manipulated. 

It is however important that the control variate is appropriately scaled. For this purpose, a parameter $\beta$ is introduced. If $\beta$ is fixed, then the estimator will be unbiased:
\begin{align}
	\mathbb{E} \big[ &\frac{1}{n} \sum_{i=1}^{n} f(x_i) - \beta \big(g(x_i) - \mathbb{E}[g(x)] \big) \big] \nonumber \\
	&= \ \mathbb{E}[f(x)] - \beta (\mathbb{E}[g(x)] -  \mathbb{E}[g(x)] ) = \mathbb{E}[ f(x)]
\end{align}

Our statistic of interest is the weighted loss. For common choices of loss functions, such as quadratic, logistic or hinge, the importance weights will generally correlate well with the weighted loss. The expected value of the importance weights is always known, since
 \begin{align}
 	\mathbb{E}_{\cal S} [ w(x) ] = \int_{\cal X} \frac{p_{\cal T}(x)}{p_{\cal S}(x)} p_{\cal S}(x) \mathrm{d} x =  \int_{\cal X} p_{\cal T}(x) \mathrm{d} x = 1 \, .
 \end{align} 
This means the weights can \emph{always} be introduced as a control variate to an importance-weighted estimator. The controlled importance-weighted risk estimator becomes:
\begin{align}
	\hat{R}_{\beta}(h_{\theta}) \! = \! \frac{1}{n} \sum_{i=1}^{n} \ell( h_{\theta}(x_i ) ,y_i) w(x_i) \! - \! \beta (w(x_i) -1) \, .
\end{align}
Note that the weights $w(x_i)$ have already been computed, implying that additional computational cost is restricted to $\beta$. The effect of the control variate on the sampling variance of the importance-weighted risk estimator can be seen via \cite{nelson1987control}:
\begin{align}
	\mathbb{V}_{\cal S} &[ \hat{R}_{\beta} ] = \mathbb{E}_{\cal S} \big[ (\hat{R}_{\beta} - R_{\beta})^2 \big] \nonumber \\
	=& \frac{1}{n} \ \mathbb{E}_{\cal S} \big[ \big( \ell_{w}(x, y) - \beta(w(x) -1) - R_{\cal W} \big)^2 \big]  \nonumber \\
	%
	%
	%
	=& \frac{1}{n} \big( \sigma^2_{\cal W} \! - \! 2\beta \mathbb{C}_{\cal S} [ \ell_w(x, \! y), \! w(x)] \! + \! \beta^2 \mathbb{V}_{\cal S} [w(x)] \big) \! = \! \frac{\sigma^2_{\beta}}{n} \label{eq:VsRb} \, ,
\end{align}
where $\ell_{w}(x,y) = \ell(h_{\theta}(x), y) w(x)$ is shorthand for the weighted loss function given a fixed classifier $h_{\theta}$, and $\mathbb{C}_{\cal S}$ refers to covariance. We can minimize this sampling variance for the scaling parameter $\beta$. Taking the derivative of (\ref{eq:VsRb}) with respect to $\beta$ and setting it to $0$ yields:
\begin{align}
	 \beta^{*} = \ \mathbb{C}_{\cal S} \big[ \ell_w(x,y), w(x) \big] \ / \ \mathbb{V}_{\cal S} \big[w(x) \big] \, . \label{eq:beta_hat}
\end{align}
Plugging $\beta^{*}$ back in simplifies the sampling variance to:
\begin{align}
	%
	%
\sigma^2_{\beta} = \sigma^2_{\cal W} \! - \! \mathbb{C}_{\cal S} \big[ \ell_{w}(x,y), w(x) \big]^2 \! / \mathbb{V}_{\cal S} \big[w(x) \big] \, .  
\end{align}
Considering that both the squared covariance term and the variance term are non-negative, the sampling variance of $\hat{R}_{\beta}$ is never larger than that of $\hat{R}_{\cal W}$ \cite{owen2000safe}. In particular, $\sigma^2_{\beta}$ can be alternatively formulated as 
$\sigma^2_{\cal W}(1 - \rho^2)$, where $\rho$ denotes the correlation between the weighted loss and the weights \cite{mcbook}. Essentially, the more the weights correlate -- \emph{positively} or \emph{negatively} -- with the weighted loss, the larger the reduction in variance.

\begin{figure}[htb]
\centering
	\includegraphics[width=.45\textwidth]{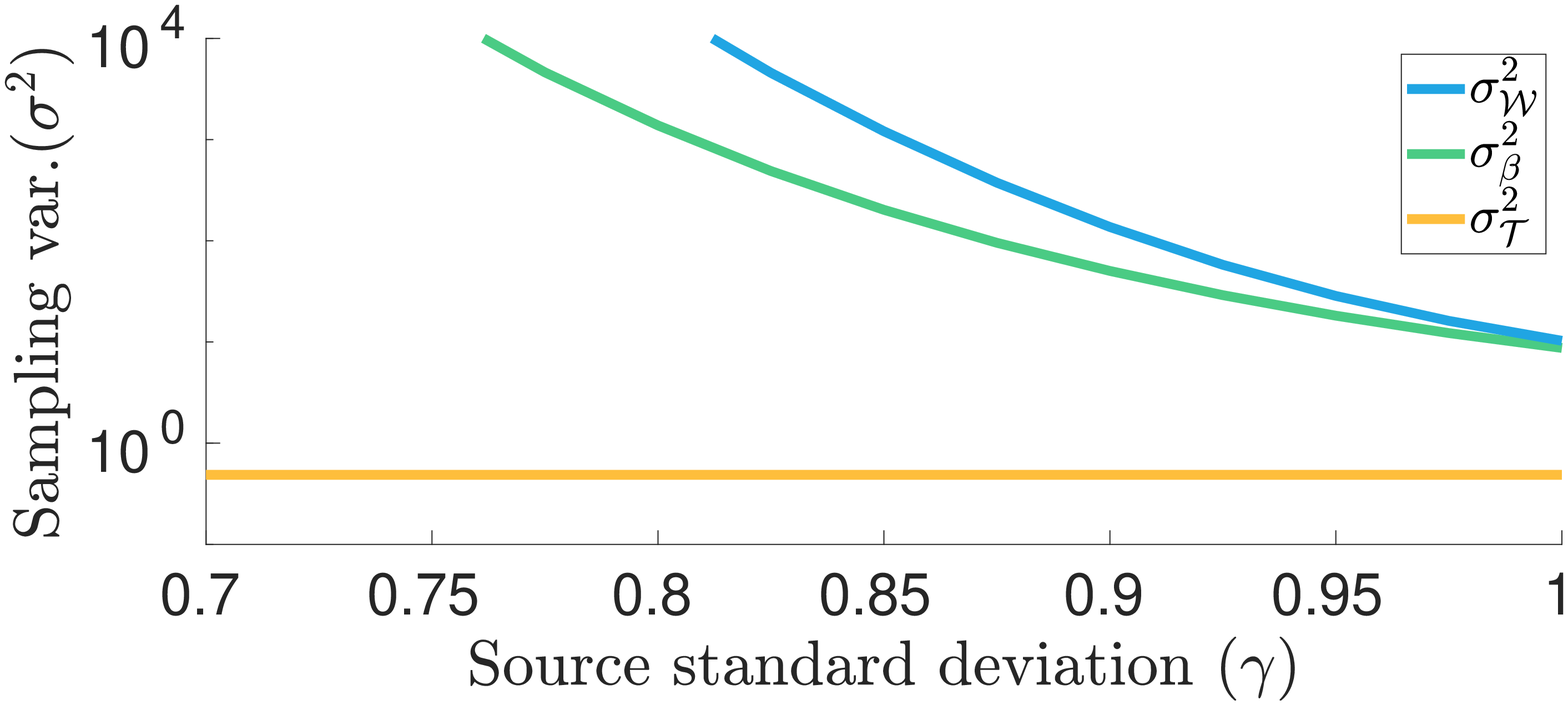}
	\caption{Sampling variance of the target (yellow, $\sigma^2_{\cal T}$), the importance-weighted (blue, $\sigma^2_{\cal W}$) and the controlled importance-weighted (green, $\sigma^2_{\cal \beta}$) risk estimators as a function of $\gamma$ in the example setting.}
	\label{fig:sv02}
\end{figure}

\noindent We computed $\sigma^2_{\beta}$ for the example setting and show it alongside $\sigma^2_{\cal W}$ and $\sigma^2_{\cal T}$ in Figure \ref{fig:sv02}. Note that $\sigma^2_{\beta}$ also diverges as $\gamma$ shrinks, like $\sigma^2_{\cal W}$, but does so at a slower rate.

\subsection{Regression estimator}
In practice, $\beta$ will need to be estimated. Both the weight variance and the covariance between the weighted loss and the weights can be estimated from data. In that case, Equation \ref{eq:beta_hat} becomes the solution to a least-squares problem \cite{mcbook}:
\begin{align}
	\hat{\beta} \! = \! \frac{\sum_{i=1}^{n} \big( \ell_w(x_i,\! y_i) \!  - \!  \sum_{i=1}^{n} \ell_w(x_i, \! y_i) \big) \big( w(x_i) \! - \! 1 \big) }{ \sum_{i=1}^{n} \big(w(x_i) -1 \big)^2} \, . \label{eq:beta_est}
\end{align}
However, $\beta$ now depends on the same observed variable as the weighted loss and the weights, which is somewhat problematic. It can be shown that the deviation of $\hat{\beta}$ from $\beta^{*}$ drops off on the order of $O(n^{-1/2})$ (see Theorem 1 from \cite{owen2000safe}). This estimation error in $\beta$ causes a bias in the risk estimator on the order of $O(n^{-1})$ \cite{owen2000safe}. 

\section{Importance-weighted cross-validation} \label{sec:cross}
Standard cross-validation will not yield optimal regularization parameter estimates under sample selection bias. Importance-weighted cross-validation is a potential solution \cite{sugiyama2007covariate}, but is not robust to large weights. The following subsections describe experiments that compare risk estimators during $k$-fold cross-validation, and evaluate their ability to find appropriate regularization parameters.

\subsection{Data}
\begin{figure}[htb]
\centering
\includegraphics[width=.2\textwidth]{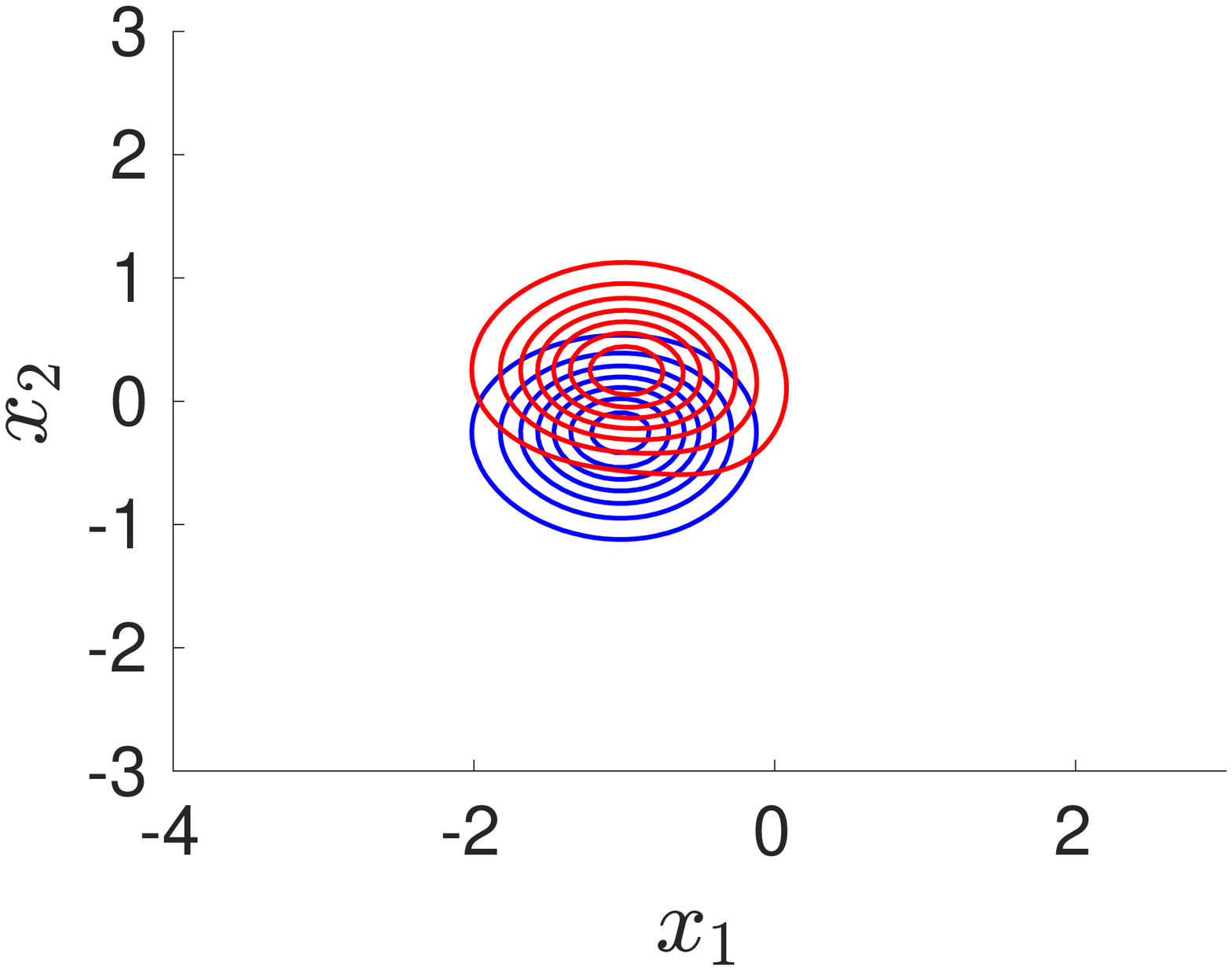} \
\includegraphics[width=.2\textwidth]{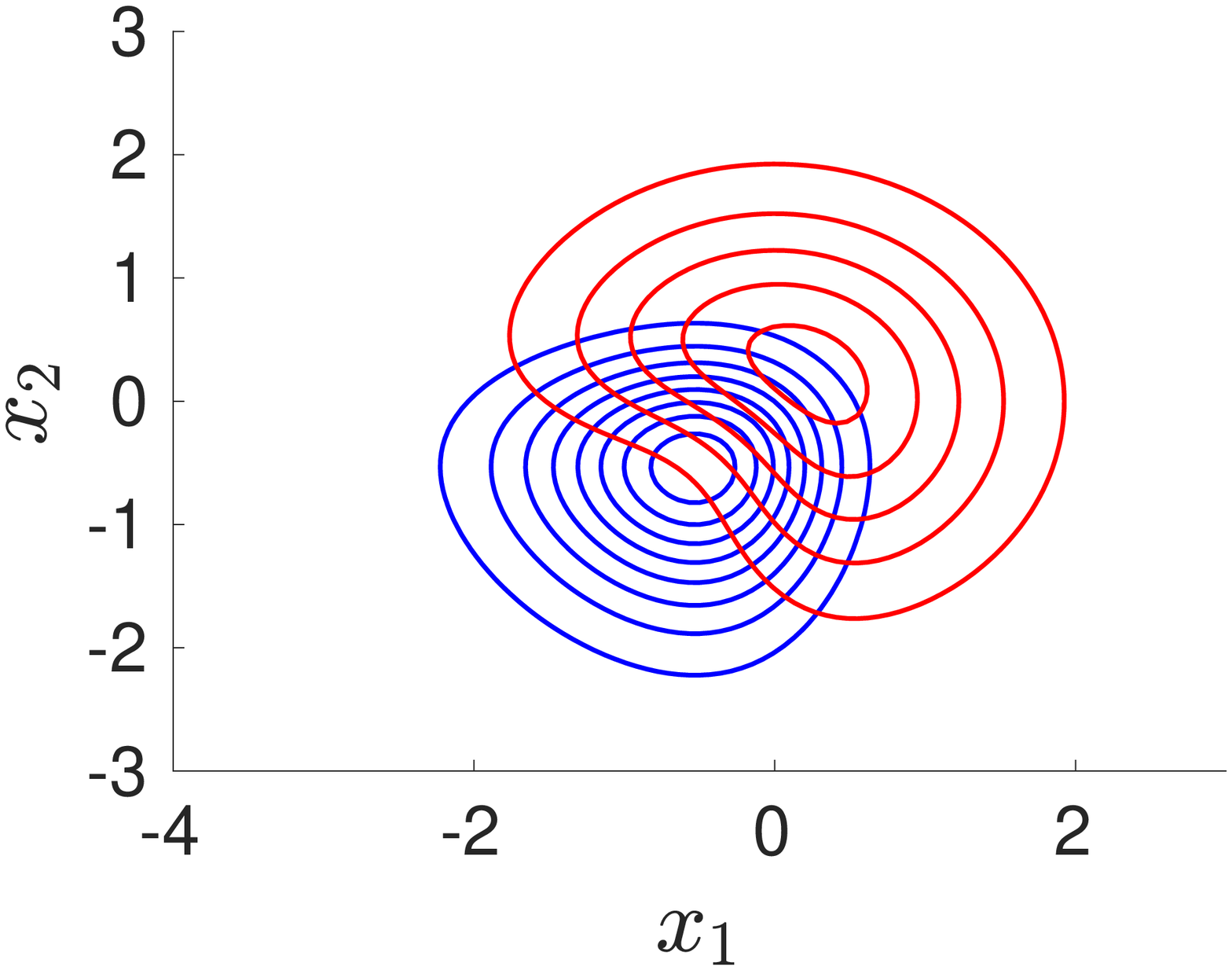} 
\caption{Synthetic problem, with $\gamma = 1/\sqrt{2}$. Class-conditional distributions for source (left) and target domain (right).}
\label{fig:problem_dists}
\end{figure}

We consider a synthetic and a natural data set. In the synthetic setting, the target data distribution is a unit bivariate normal $\mathcal{N}([0, 0] \given I)$, while the source data distribution consists of a bivariate normal with a shifted mean and a scaled identity covariance matrix, $\mathcal{N}([-1, 0] \given \gamma I)$. The posterior distribution is of the form: $p(y=-1 \mid x_1, x_2) = 1 - \Phi(-[x_1, x_2])$ and $p(y=1 \mid x_1, x_2) = \Phi(-[x_1, x_2])$. Figure \ref{fig:problem_dists} visualizes the class-conditional distributions $p(x \given y)$ for $\gamma = 1 / \sqrt{2}$. Note that the source domain is a local sampling of the larger target domain and that the nonlinear nature of the underlying decision boundary is not apparent. We draw data sets using rejection sampling, with $50$ source samples and $1000$ target samples.

The natural setting is derived from the ozone level detection data set from the UCI machine learning repository\footnote{\url{https://archive.ics.uci.edu/ml/datasets}}. The task is to predict ozone days versus normal weather days, given various weather station measurements. All samples with missing values are removed, the data has been z-scored, and it has been projected onto its first 10 principal components. The source domain is a local sampling in time: a Gaussian centered at the start date with a standard deviation that is a proportion $\gamma$ of the total number of samples. For small values of $\gamma$, the source domain contains only samples close to the start date. For large values, the source domain is roughly equally spread over time. We draw 80 samples without replacement from both classes to form the source data set. These $80$ samples are also included in the target data, so as not to form a 'hole' in feature space.

For each setting, we perform 100 000 repetitions of sampling a data set. The top 10\% of repetitions with the largest weight variance are considered the data sets with problematically large weights. 

\subsection{Experimental setup}
The source set is split into $k=5$ folds, a classifier is trained using a particular value for the regularization parameter $\lambda$ and its loss is computed on each data point in the held-out validation set. That loss is stored, along with those points' estimated weights. An importance-weighted $L_2$-regularized linear least-squares classifier is taken. Its parameters are estimated through $\theta_{\lambda} = (X_{t} W_t X_{t}^{\top} + \lambda I)^{-1}(X_{t}^{\top} W_t y_{t})$ where $t$ indicates training indices and $W$ is a diagonal matrix with the estimated weights as its entries. For $\lambda$, we considered a range from $10^{-3}$ to $10^6$ over 200 logarithmic steps. 

Weights are estimated by fitting a normal distribution to data from each domain, and computing the ratio of the target probability over the source probability of each source data point $\hat{w}(x_i) = \hat{p}_{\cal T}(x_i) / \hat{p}_{\cal S}(x_i)$. We compare the importance-weighted risk estimator ($\hat{R}_{\hat{w}}$) with its control variate counterpart ($\hat{R}_{\hat{\beta}}$). We average their estimated risks over all data sets and specifically over the 10\% of data sets with the largest weight variance (indicated with "$>$" in the legend of Figures \ref{fig:2DG_varW10}, \ref{fig:ozone_varW10} and \ref {fig:2DG_iwe}). We also include validation on the labeled target samples ($\hat{R}_{\cal T}$) as the oracle solution. 
After risk estimation, the $\lambda$ is selected that minimized risk. The classifier is then re-trained using all source data and the selected $\lambda$, and evaluated using the target risk based on the true target labels as the final measure. This process is repeated for each data set and we report the final average as $\bar{R}_{\cal T}$. 

Repeating the above procedure for each data set allows us to perform non-parametric hypothesis tests for statistically significant differences between the final risks of the estimators. Since the data is paired and the estimators' sampling distributions are skewed beyond normality, we employ a Wilcoxon signed-rank test \cite{conover1980practical}.

\subsection{Results}
\begin{figure}[htb]
\includegraphics[width=.44\textwidth,center]{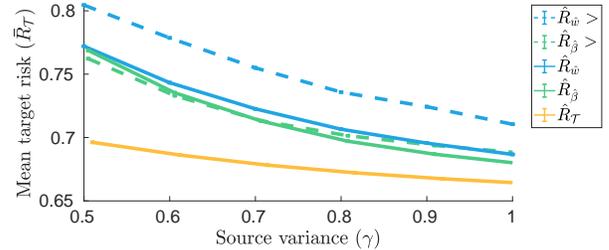}
\caption{Mean target risks for the synthetic problem setting, as a function of source variance $\gamma$.}
\label{fig:2DG_varW10}
\end{figure}

Figure \ref{fig:2DG_varW10} presents the final target risks for each risk estimator on the synthetic problem setting. Note that the errorbars in the plot are too small to see. When looking at all data sets, there is nearly no difference between $\hat{R}_{\hat{w}}$ and $\hat{R}_{\hat{\beta}}$. This indicates that the addition of the control variate does not deteriorate importance-weighted cross-validation in general. When looking at the sets with large weight variance, it can be seen that $\hat{R}_{\hat{w}}$ deteriorates strongly, while $\hat{R}_{\hat{\beta}}$ does not. This shows that the addition of the control variate makes the importance-weighted risk estimator more robust to large weights. The difference between $\hat{R}_{\hat{w}}$ and $\hat{R}_{\hat{\beta}}$ for the data sets with large weight variance (dotted lines) is statistically significant for each value of $\gamma$, with the largest $p$-value on the order of $10^{-30}$.

\begin{figure}[htb]
\includegraphics[width=.44\textwidth,center]{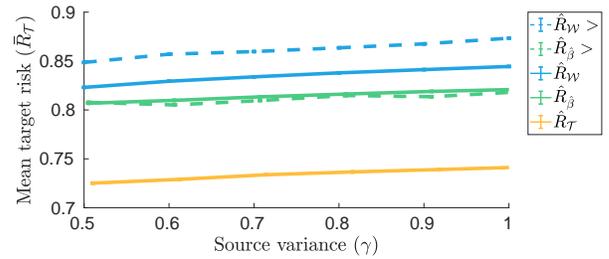}
\caption{Mean target risks for Ozone level detection setting, as a function of the relative scale of the source domain $\gamma$.}
\label{fig:ozone_varW10}
\end{figure}

Figure \ref{fig:ozone_varW10} presents the mean target risks in the Ozone problem setting. Note that the controlled estimator $\hat{R}_{\hat{\beta}}$ is essentially unaffected by the large weights since it produces the same final risks in both the case of all data sets and the top $10\%$ of largest weight variance. Both sets of risks are below that of the uncontrolled estimator $\hat{R}_{\hat{w}}$, which still deteriorates in the data sets with large weight variance. All differences between $\hat{R}_{\hat{w}}$ and $\hat{R}_{\hat{\beta}}$ are statistically significant for all values of $\gamma$, with the largest $p$-value on the order of $10^{-50}$. Lastly, all risks increase slightly as $\gamma$ grows, which is due to less data outliers around the start date. 

\subsection{Weight estimators}
\begin{figure}[htb]
\includegraphics[width=.44\textwidth,center]{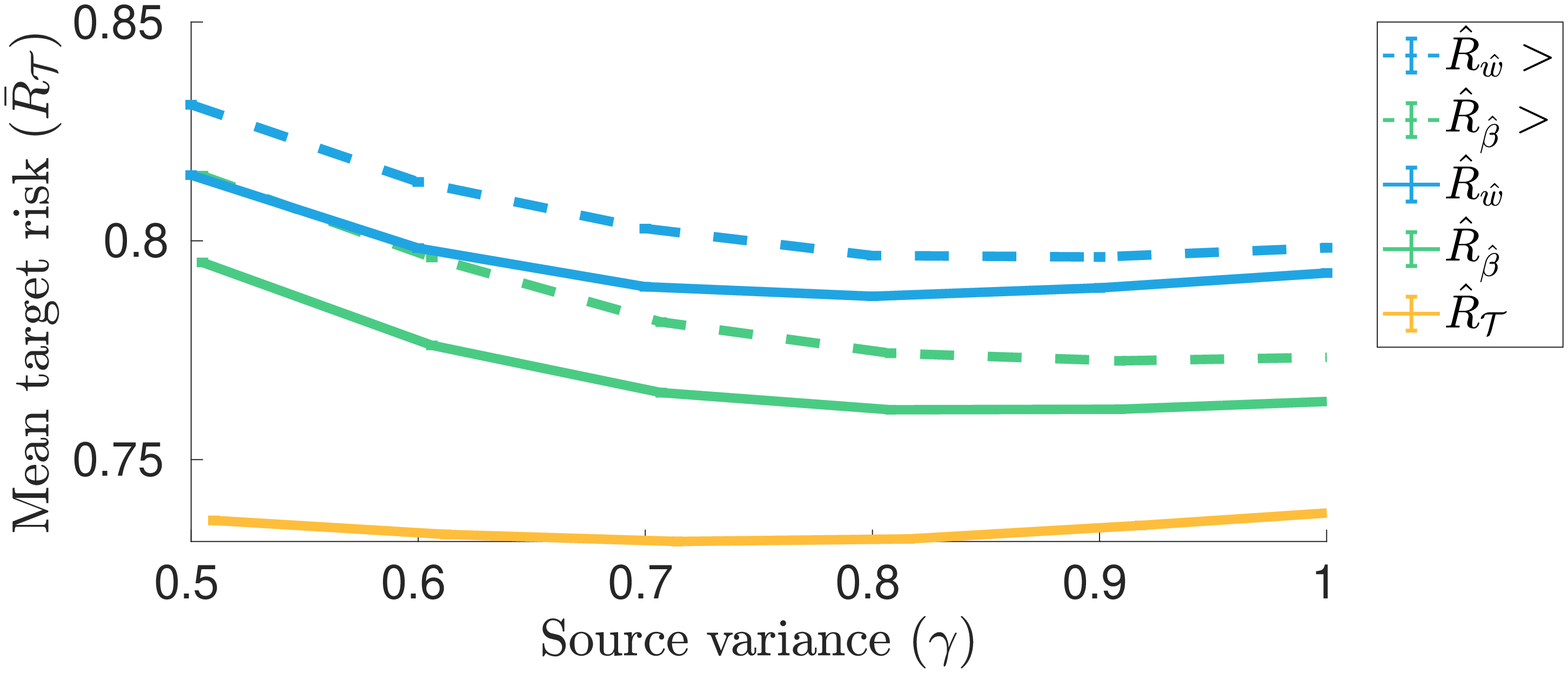} \\
\includegraphics[width=.44\textwidth,center]{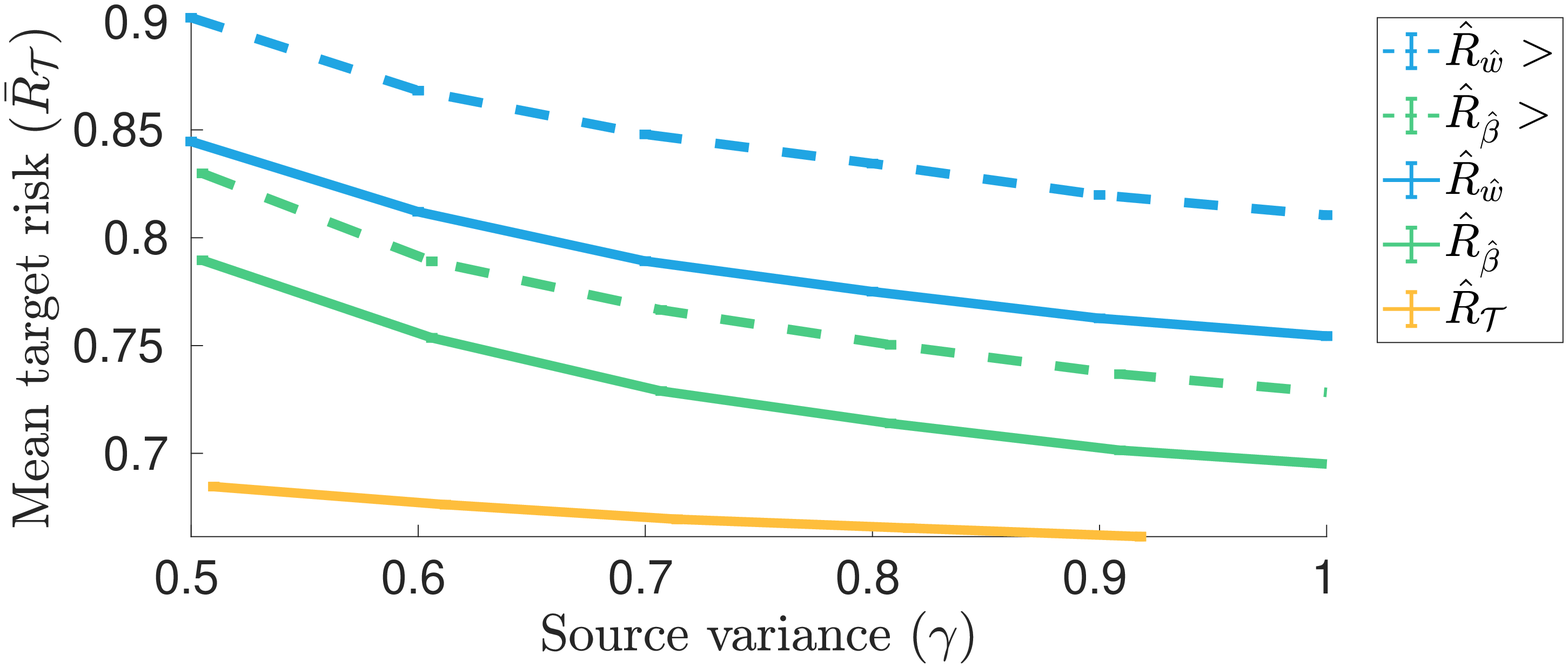} 
\caption{Mean target risks for the synthetic problem setting, as a function of $\gamma$. (Top) KMM, (bottom) KLIEP.}
\label{fig:2DG_iwe}
\end{figure}

The effect of the control variate is independent of the choice of importance-weight estimator. We illustrate this point by performing the same experiment on the synthetic setting using Kernel Mean Matching (KMM) \cite{huang2007correcting} and the Kullback-Leibler Importance Estimation Procedure (KLIEP) \cite{sugiyama2008direct}. Figure \ref{fig:2DG_iwe} shows similar results as in Figure \ref{fig:2DG_varW10}, except that the control variate even leads to better final target risks for the average over all data sets (green lines are below blue lines). 

\noindent Some non-parametric weight estimators are formulated such that they naturally avoid large weights. However, this behavior depends on one or more hyperparameters, such as a kernel bandwidth parameter. Unfortunately, finding an optimal bandwidth parameter would require cross-validation, and consequently, such estimators are not suited to cross-validation procedures. One has to resort to heuristics, such as Silverman's rule or the average distance to $k$-nearest-neighbours. In the above experiment, we set the kernel bandwidth in KMM and KLIEP to the average Euclidean distance from the source points to their five nearest neighbours of the target points. 

\section{Conclusion \& Discussion} \label{sec:discussion}
We introduced a control variate to reduce the sampling variance of the importance-weighted risk estimator. With its inclusion, the importance-weighted risk estimator is more robust large weight variance. Consequently, during $k$-fold cross-validation, it selects better hyperparameters than the uncontrolled importance-weighted risk estimator.

We have studied an additive linear control variate. Alternatively, one could consider more complex control variates, such as higher-order moments of the weights, or multiplicative control variates \cite{mcbook}. If these have a stronger correlation with the weighted loss, they could lead to larger reductions in variance. However, there are a wide variety of possible alternatives, and it is unclear how to search over these.

\bibliography{kouw_mlsp19a_short}
\bibliographystyle{IEEEbib}

\end{document}